\documentclass{llncs}

\usepackage{amssymb,amsmath,xcolor,url,underscore}

\usepackage{graphicx}
\usepackage{epsfig,color}

\usepackage{hyperref}
\usepackage{listings}
\usepackage[space]{grffile}

\graphicspath{{./figs/}}
\usepackage{epstopdf}
\usepackage{float}

\title{Style Imitation and Chord Invention in Polyphonic Music with Exponential Families}

\author{Ga\"etan Hadjeres\inst{1, 2}  \and Jason Sakellariou\inst{1,2} \and Fran\c{c}ois Pachet\inst{1}  }

\institute{Sony Computer Science Laboratories, Paris
\and LIP6, Université Pierre et Marie Curie
\email{\\gaetan.hadjeres@etu-upmc.fr, js.sakel@gmail.com, pachetcsl@gmail.com }}

\begin{document}
\maketitle

\begin{abstract}
Modeling polyphonic music is a particularly challenging task because of the intricate interplay between melody and harmony. 
A good model should satisfy three requirements: statistical accuracy
(capturing faithfully the statistics of correlations at various
ranges, horizontally and vertically), flexibility (coping with
arbitrary user constraints), and generalization capacity (inventing
new material, while staying in the style of the training
corpus). Models proposed so far fail on at least one of these
requirements.
We propose a statistical model of  polyphonic music, based on the maximum entropy principle. This model is able to learn and reproduce pairwise statistics between neighboring note events in a given corpus. The model is also able to invent new chords and to harmonize unknown melodies. We evaluate the invention capacity of the model by assessing the amount of cited, re-discovered, and invented chords on a corpus of Bach chorales. We discuss how the model enables the user to specify and enforce user-defined constraints, which makes it useful for style-based, interactive music generation.
\end{abstract}

\section{Introduction}
\label{intro}

Polyphonic tonal music is often considered as a highlight of Western civilization. Today's music is still largely based on complex structures invented and developed since the Renaissance period, and modeled, e.g. by Jean Philippe Rameau \cite{Rameau} in the XVII$^{th}$ century.
In particular, polyphonic music is characterized by an intricate
interplay between melody (single-voice stream of notes) and harmony
(progression of simultaneously-heard notes). 
Additionally, composers tend to develop a specific \textit{style}, that influences the way notes are combined together to form a musical piece. 

Many models of polyphonic music have been proposed since the 50s (see \cite{fernandez2013ai} for a comprehensive survey), starting with the famous Illiac Suite, which used Markov chains to produce 4-voice music, controlled by hand-made rules \cite{LHiller}. 
In this paper we address the issue of learning agnostically the style of a polyphonic composer, with the aim of producing new musical pieces, that satisfy additional user constraints.

In practice, a good model should satisfy three requirements: statistical \textit{accuracy} (capturing faithfully statistics of correlations at various ranges, horizontally and vertically), \textit{flexibility} (coping with arbitrary user constraints), and \textit{generalization} capacity (inventing new material, while staying in the style of the training corpus). 
Models proposed so far fail on at least one of these requirements. 
\cite{eppe2015computational} propose a chord invention framework but is not based on agnostic learning, and requires a hand-made ontology.
The approach described in \cite{pachet:14a} consists in a dynamic
programming template enriched by constrained Markov chains. This
approach generates musically convincing results \cite{pachet:16a} but is \textit{ad hoc} and specialized for jazz. Furthermore it does not invent any new voicing by construction (the vertical ordering of the notes in a chord).
\cite{hild1992harmonet} and \cite{allan2005harmonising} describe a HMM approach trained on an annotated corpus. This model imitates the style of Bach chorales,
as shown by cross entropy measures. However, it is also not able to produce new voicings by construction, and only replicates voicings found in the training corpus.
Another related approach is \cite{kaliakatsos2014probabilistic} which uses HMMs on chord representations (based on an expert knowledge of the common-practice harmony) called General Chord Type (GCT) \cite{cambouropoulos2014idiom} to generate homorhythmic sequences. Those models are not agnostic in the sense that they include a priori knowledge about music such as the concept of dissonance, consonance, tonality or scale degrees. 
Agnostic approaches using neural networks have been investigated with
promising results. In \cite{boulanger2012modeling}, chords are modeled with Restricted Boltzmann Machines (RBMs).  Their temporal dependencies are learned using Recurrent Neural networks (RNNs).
Variations of these architectures have been developed, based on Long Short-Term Memory (LSTM) units \cite{Lyu2015} or GRU (Gated Recurrent Units) \cite{chung2014empirical}. 
However, these models require large and coherent training sets which are not always available. More importantly, it is not clear how to enforce additional user constraints (flexibility). Moreover, their invention capacity is not demonstrated.

In this paper we introduce a graphical model based on the maximum entropy principle for learning and generating  polyphony. Such models have been used for music retrieval applications \cite{pickens2005markov}, but never, to our knowledge, for polyphonic music generation. 

This model requires no expert knowledge about music and can be trained on small corpora. Moreover, generation is extremely fast. 

We show that this model can capture and reproduce pairwise statistics at possibly long range, both horizontally and vertically. These pairwise statistics are also able, to some extent, to capture implicitly higher order correlations, such as the structure of 4-note chords.
The model is flexible, as it allows the user to post arbitrary unary constraints on any voice. Most importantly, we show that this model exhibits a remarkable chord invention capacity. In particular we show that it produces harmonically consistent sequences using chords which did not appear in the original corpus. 

In Sect.~\ref{model} we present the model for $n$-parts  polyphony generation. 
In Sect.~\ref{discovery}, we report experimental results about chord invention. 
In Section~\ref{sec:reharmonization} we discuss a range of interactive applications in music generation. 
Finally, we discuss how the ``musical interest'' of the generated sequences depends on the choice of
our model's hyperparameters in Sect.~\ref{parameterChoice}.

\section{The model}
The model we propose is based on a maximum entropy model described in \cite{sakellariou:15a}. This model is extended to handle several voices instead of one, and to establish vertical as well as diagonal interactions between notes. We formulate the model as an exponential family obtained by a product of experts (one for each voice). 
 
\label{model}
\subsection{Description of the model}
\label{modelDescription}
We aim to learn sequences $s$ of $n$-part chord sequences. A sequence $s = [c_1, \dots, c_l]$ is composed of $l$ chords where the $j$th chord is denoted
\[ c_j := [s_{1j}, s_{2j}, \dots, s_{nj}],\] with note $s_{ij}$
considered as an integer pitch belonging to the pitch range
$\mathcal{A}_i \subset \mathbf{Z}$. The $i$th part or \emph{voice}
corresponds to \[v_i := [s_{i1}, s_{i2},  \dots, s_{il}].\]

Our model is based on the idea that chord progressions can be
generated by replicating the occurrences of pairs of neighboring
notes. It is invariant by  translation  in time, it aims at capturing the local ``texture'' of the chord sequences. Similar ideas have been shown to be successful in modeling highly combinatorial and arbitrary structures such as English
four-letter words \cite{stephens2007toward}. 

\begin{figure}

    \centering
  \includegraphics[height=3cm, width=4cm]{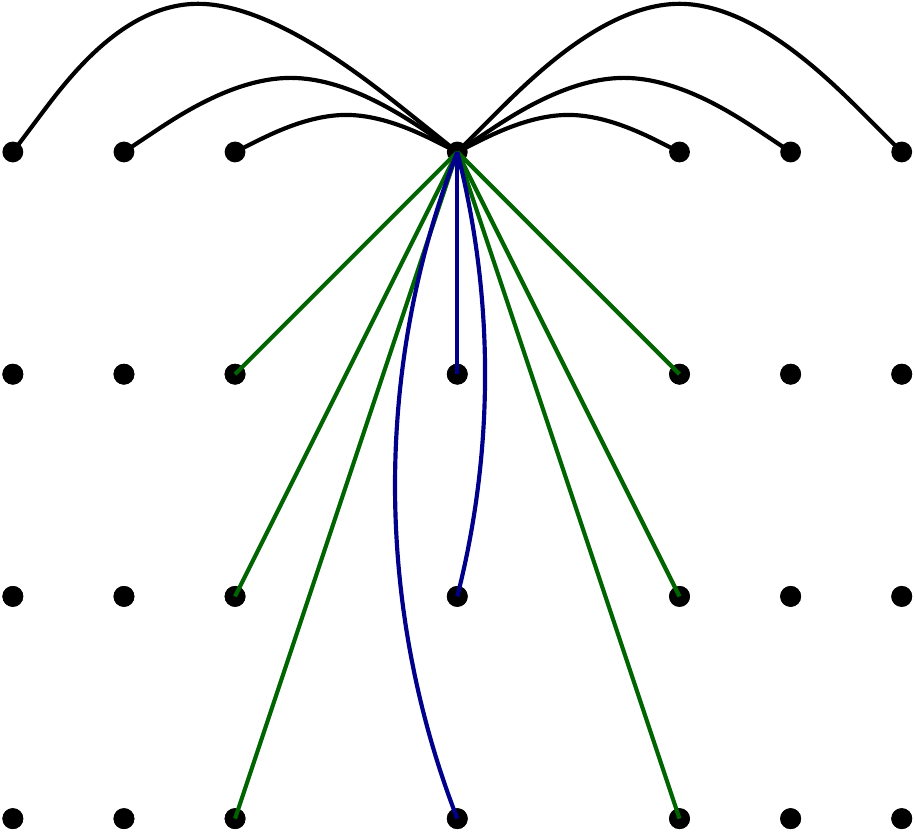} \quad
  \includegraphics[height=3cm, width=4cm]{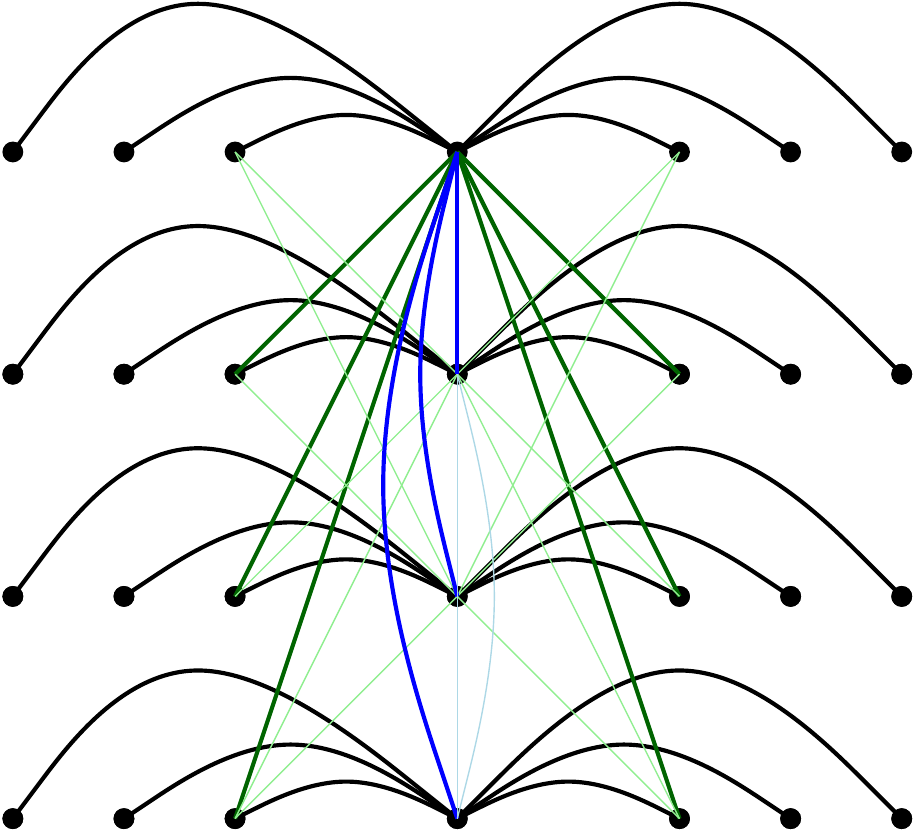}
  \caption{Example of binary connections involving  the first voice's
    fourth note (on the left) and the fourth chord (on the
    right). Horizontal connections are in black, vertical in green and
    diagonal in blue.}
\label{graphicalModel}
  \end{figure}

We denote
by  $K$ the model \emph{scope}, which means that we consider that chords
distant by more than $K$ time steps are conditionally independent
given all other variables. We focus on the interaction between neighbouring notes and try to replicat the co-occurrences notes. A natural way to formalize this is to introduce a family of functions (or features) such that each member of this family counts the number of occurrence of a given pair of notes. As 
a result, the finite number of  features we want to  learn can be written as a family
\begin{equation}
  \label{eq:4}
\left\{ f_{ab,ijk}  \quad \mathrm{s.t.}  \quad
  \begin{tabular}{l}
    $a \in \mathcal{A}_i, \quad    b \in \mathcal{A}_j $\\
    $i, j \in [1,n], \quad    k \in [-K,K]$
  \end{tabular}
\right\}
\end{equation}
of functions over chord sequences where
\begin{equation}
  \label{eq:5}
 f_{ab,ijk}(s) := \sharp \left\{ m \quad \mathrm{s.t.} \quad s_{i,m} =
   a \ \mathrm{ and } \ s_{j,m+k} = b \right\} 
\end{equation}
stands for the number of occurrences of pairs of notes $(a,b)$ such
that note $a$ at voice $i$ precedes by $k$ time steps note $b$ at
voice $j$ in the chord sequence $s$.  We can represent this family of
\emph{binary connections} as
a graphical model as can be seen in Fig.~\ref{graphicalModel} where each
subfamily \[\{f_{ab,ijk}, \quad \forall a \in \mathcal{A}_i, \quad    b \in
\mathcal{A}_j\} \] is represented as a link between two notes. 
Our model has also \emph{unary} parameters, acting on single notes and modeling
the single note marginal distributions. For notational convenience we will treat 
these unary parameters as binary
connections between pairs of identical notes (connections such that
$a{=}b$, $i{=}j$, $k{=}0$) and call them \emph{local fields} after the 
corresponding statistical physics terminology. 
We differentiate four types of connections: unary (local fields) and binary
``horizontal'', ``vertical'' and ``diagonal'' connections.  We implicitly identify $f_{ab,ijk}$ 
with $f_{ba, ji(-k)}$, for all $k \in  [-K,K]$, $i, j \in [1,n]$.

From now on, we will denote the set of indexes of the family $\{f_{ab,ijk}\}$ by $\mathcal{P}$. We note that there is approximately
\begin{equation}
  \label{eq:numPars}
 n^2 (2K + 1)\left| {\mathcal{A}}\right|^2
\end{equation}
indexes,  where $|\mathcal{A}|$ stands for the mean alphabet size 
\[|\mathcal{A}| = \frac{1}{n} \sum_{i=1}^n |\mathcal{A}_i|.\] Using
only a subset of the family $\{f_{ab,ijk}\}$ given by (\ref{eq:4}) can reduce the number of
parameters while leading to good results. Indeed, if we consider that
notes  in different voices are conditionally independent if they are
distant by more than  $L \leq K$ time steps, we obtain a index set of
size approximately equal to 
\[ (n \times K + \frac{n^2 - n}{2} L)|\mathcal{A}|^2.\]
In the following, $\mathcal{P}$ can designate the whole set of
indexes within scope $K$ as well as any of its subset.

Let $\mu^{ab,ijk}$ for $(ab,ijk) \in \mathcal{P}$ be real
numbers. From all distributions $P$ such that the averages over all
possible sequences of length $l$ verify
\begin{equation}\label{means} \sum_s P(s) f_{ab,ijk}(s) = \mu^{ab,ijk}
  \quad \forall (ab,ijk) \in \mathcal{P},
\end{equation}
it is known that the exponential distribution  with
statistics $\{f_{ab,ijk}\}$ and parameters $\{\mu^{ab,ijk}\}$ is the
one of maximum entropy (i.e. which makes the least amount of
assumptions) satisfying Eq.~(\ref{means}). We thus consider
an energy-based model of parameter \[\theta := \left\{
\theta^{ab,ijk} \in \mathbf{R} \quad \forall (ab,ijk) \in
\mathcal{P} \right\}\]
given by 
\begin{equation}
  \label{eq:6}
 P(s | \theta) = \frac{e^{-E(s, \theta)}}{Z(\theta)},
\end{equation}
where
\begin{equation}
  \label{eq:7}
E(s, \theta) := -\sum_{ab,ijk} \theta^{ab,ijk} f_{ab,ijk}(s)
\end{equation}
is usually called the \emph{energy} of the sequence $s$ and  \[Z(\theta) =
\log(\sum_{s}  e^{  - E(s, \theta)})\] the \emph{normalizer} or
\emph{partition function} such that $P$ defines a probability function
over sequences of size $l$. The sum in the partition function is for
every $s$ of size $l$. There are approximately $|A|^l$ such
sequences, which make the exact computation of the partition function
intractable in general.


\subsection{Training}
 We are given a training dataset $\mathcal{D}$ composed of $N$ $n$-part sequences
$s^{(1)},\dots, s^{(N)}$ that we suppose, for clarity, to be
concatenated in one sequence $s$. 
Since we are dealing with discrete data, gradient techniques such as
score matching \cite{hyvarinen2005estimation} cannot be used. Instead, we choose
to minimize  the negative \emph{pseudo-log-likelihood}
\cite{ravikumar2010high,ekeberg2013improved} of the data
to find an approximation of the true maximum likelihood estimator. It
consists in approximating the negative log-likelihood function
\begin{equation}
  \label{eq:loglikelihood}
\mathcal{L}(\theta, s) = - \log P(s |
\theta)
\end{equation}
 by the mean of conditional log-likelihoods of a note given the others, that is 
\begin{equation}
  \label{eq:1}
 \mathcal{L}(\theta, s) = - \frac{1}{nl}\sum_{ij} \log P(s_{ij} | s \setminus
 s_{ij}, \theta),
\end{equation}
where $s \setminus s_{ij}$ denotes all notes in $s$ except $s_{ij}$. The conditional probabilities are calculated as 
\begin{equation}
  \label{eq:10}
P(s_{ij} | s \setminus s_{ij}, \theta) =\frac{P(s,\theta)}{\sum_{\substack{c \in \mathcal{A}_i\\
s' \setminus s'_{ij}= s \setminus s_{ij} \\
s_{ij} = c}}P(s',\theta)} 
\end{equation}
where the sum in the denominator is on chord sequences $s'$ equal to $s$ except for the note in position $ij$.

Due to the particular structure of the probability density function
(\ref{eq:6}) and the choice of the statistics (\ref{eq:4}),
we note that we can write 
\[P(s_{ij} | s \setminus
 s_{ij}, \theta) = P(s_{ij} | \mathcal{N}_K(i,j,s), \theta),\] where $\mathcal{N}_K(i, j, s)$ stands for the neighbors of note
$s_{ij}$ in $s$ which are at a distance inferior to $K$ time steps. We now express our dataset $\mathcal{D}$ as a
set of samples $(x,\mathcal{N})$ consisting of a note $x$ and its
$K$-distant neighbors, ignoring border terms whose effect is negligible. More precisely, we write the dataset
\begin{equation}
\begin{split}
  \label{eq:12}
\mathcal{D}  = \{ \left(s_{ij},\mathcal{N}_K(i,j,s)\right), &
  \\  \forall i \in [1, n] & , j \in [K + 1, l-K -1] \}
\end{split}
\end{equation}
and split it
into $n$ datasets $\mathcal{D}_i$ such as
\[ \mathcal{D}_i = \left\{ \left(s_{ij},  \mathcal{N}_K(i,j,s)\right),
  \quad  j \in [K + 1, l - K - 1] \right\}.\] Each element in this dataset can
still be
considered as the subsequence it comes from also written $(x, \mathcal{N})$.
Those notations set, we can rewrite  Eq.~\ref{eq:1} as
\begin{equation}
  \label{eq:8}
\begin{split}
 \mathcal{L}(\theta, \mathcal{D}) &  = - \frac{1}{\sharp \mathcal{D}} \sum_{(x,\mathcal{N}) \in \mathcal{D}}
 \log P(x |\mathcal{N}, \theta) \\ &:=   \frac{1}{n}  \sum_{i = 1}^n 
 \mathcal{L}_i(\theta,\mathcal{D}_i),
\end{split}
\end{equation}
where  \[\mathcal{L}_i(\theta, \mathcal{D}_i) = - \frac{1}{\sharp \mathcal{D}_i}
\sum_{(x,\mathcal{N}) \in \mathcal{D}_i} \log P(x | \mathcal{N}, \theta) \] is the negative conditional log-likelihood function for voice $i$.
This consists in minimizing the mean of $n$ negative log-likelihood
functions $\mathcal{L}_i$ (one for each voice) over the data $\mathcal{D}_i$.
This method has the advantage of being tractable since there are only
$\sharp \mathcal{A}_i$ terms in the denominator of Eq.~\ref{eq:10} and
can lead to good estimates  \cite{arnold1991pseudolikelihood}. This
can be seen as the likelihood of a product of experts using $n$
modified (addition of vertical and diagonal connections) copies of the model presented in \cite{sakellariou:15a}.

We need to find the parameters minimizing the sum of $n$ convex
functions. Computing the gradient of $\mathcal{L}_i$ for $i \in [1,n]$
with respect to any parameter $\theta_{*}$, $* \in \mathcal{P}$,  gives us
\begin{equation}
  \label{eq:3}
\begin{split}
 \frac{\partial \mathcal{L}_i(\theta, \mathcal{D}_i)}{\partial
   \theta_{*}} =  & \\  \frac{1}{\sharp \mathcal{D}_i}
 \sum_{\substack{s = (x,\mathcal{N}) \\ (x, \mathcal{N}) \in \mathcal{D}_i}}  & \left( \frac{ f_{*}(s) e^{ -E( s,
       \theta)}}{\sum_{\substack{s' = (z, \mathcal{N}) \\ z \in \mathcal{A}_i}} f_{*}(s') e^{ -E( s', \theta)}}      -
   f_{*}(s) \right).
\end{split}
\end{equation}
This can be written as 
\[  \frac{\partial \mathcal{L}_i(\theta,\mathcal{D}_i)}{\partial
  \theta_{*}} = \langle f_{*} \rangle_{P(. | \mathcal{N}, \theta)} - \langle f_{*} \rangle_{\mathcal{D}_i} \]
  which is the difference between the average value of $f_{*}$ taken with respect to the conditional distribution (\ref{eq:10})  and its empirical value.

A preprocessing of the corpus is introduced in order to efficiently compute the gradient sums.

Finally, the function $g(\theta)$ we will optimize is the
$L1$-regularized version of $\mathcal{L}(\theta, \mathcal{D})$ with
regularization parameter $\lambda$, which means that we consider
\begin{equation}
  \label{eq:11}
  g(\theta) =  \mathcal{L}(\theta, \mathcal{D}) + \lambda \|\theta\|_1,
\end{equation}
where $\|.\|_1$ is the usual $L1$-norm, which is the sum of the
absolute values of the coordinates of the parameter. This is known as the Lasso
regularization  whose effects are widely discussed
throughout statistical learning literature \cite{friedman2001elements}.




\subsection{Generation}

Generation is performed with the
Metropolis-Hastings algorithm, which is an extensively used sampling
algorithm (see \cite{chib1995understanding} for an introduction). 
\label{generation}
Its main feature
is the possibility to sample from an unnormalized distribution since it
only needs  to compute ratios of probabilities
\begin{equation}
  \label{eq:2}
\alpha := \frac{P(s', \theta)}{P(s, \theta)} 
\end{equation}
between two sequences $s'$ and  $s$. With the specific proposals detailed below, we just need to start from a random sequence $s$ as our
current sequence and repeat the following: choose a
proposal $s'$, compute $\alpha$ and then accept as our current
sequence  $s'$ with probability $\min(\alpha, 1)$ or  reject this
proposal with probability $1 - \min(\alpha, 1)$ and keep $s$. After a
sufficient number of iterations of this procedure, we are assured that
the sequences obtained are
distributed according the objective distribution $P(. | \theta)$.

 We used
this algorithm to generate chord sequences by choosing  $s'$ uniformly
among all sequences differing by only one note from $s$. In fact, with
this algorithm, we
can also enforce unary constraints on the produced sequences.
If we only propose sequences $s'$ containing a sequence $\{n_{ij}\}_{(ij)\in \mathcal{C}}$ of imposed notes, i.e. such that
\[ s_{ij} = n_{ij}, \quad \forall (ij) \in \mathcal{C} \]
where $\mathcal{C}$ contains the indexes of the constrained notes, the
Metropolis-Hastings algorithm samples from the distribution
\[ P( s | s_{ij} = n_{ij}, \quad \forall (ij) \in \mathcal{C}).\] This
enables us to provide reharmonizations of a given melody. We can also
add  pitch range constraints on given notes, which means that we
are given a set  \[ \left\{ \mathcal{A}_{ij} \subset
\mathcal{A}_i, \quad \forall (ij) \in \mathcal{C} \right\}\]  such that we only
propose sequences $s'$ where 
\[ s_{ij} \in  \mathcal{A}_{ij}, \quad \forall (ij) \in \mathcal{C}.\]
This can be used to impose chord labels without imposing its voicing.

\section{Experimental Results}
\label{exps}
We report experiments made using a set of 343 four-voice ($n=4$) chorale harmonizations by Johann Sebastian Bach \cite{bach1985389}. In order to evaluate the chord creation capabilities, we only retained in our chord sequences  notes heard on beats. Sect.\ref{rhythm} shows how we can easily produce rhythm using our model.
We transposed every chorale in the key of C and considered 2 corpora: a corpus with chorales in a major key and a corpus with chorales in a minor key.

The quadratic number of parameters (thanks to the exclusive use of binary connections) makes the learning phase computationally tractable.

We used the L-BFGS method from Stanford CoreNLP optimization package
\cite{manning2014stanford} to perform the gradient descent.


In the next sections, we report on accuracy (the style imitation capacity of the model), its 
 invention capacity, and flexibility.

\subsection{Style imitation}
\label{imitation}
\begin{figure*}
    \label{modelVSoriginalConnections}
    \centering
    \includegraphics[trim = 0cm 4cm 0cm 4.5cm, clip,
    height=4cm]{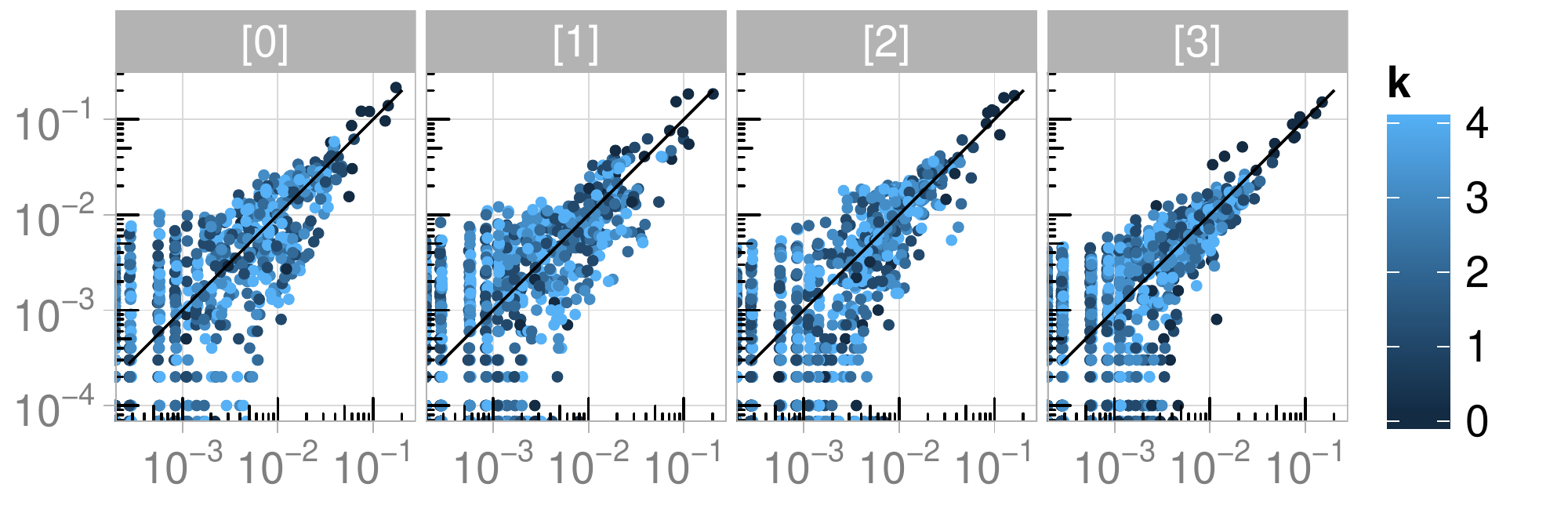}

\vspace{-0.8cm}

    \includegraphics[trim = 0cm 5cm 0.5cm 6cm, clip,
    height=4.5cm]{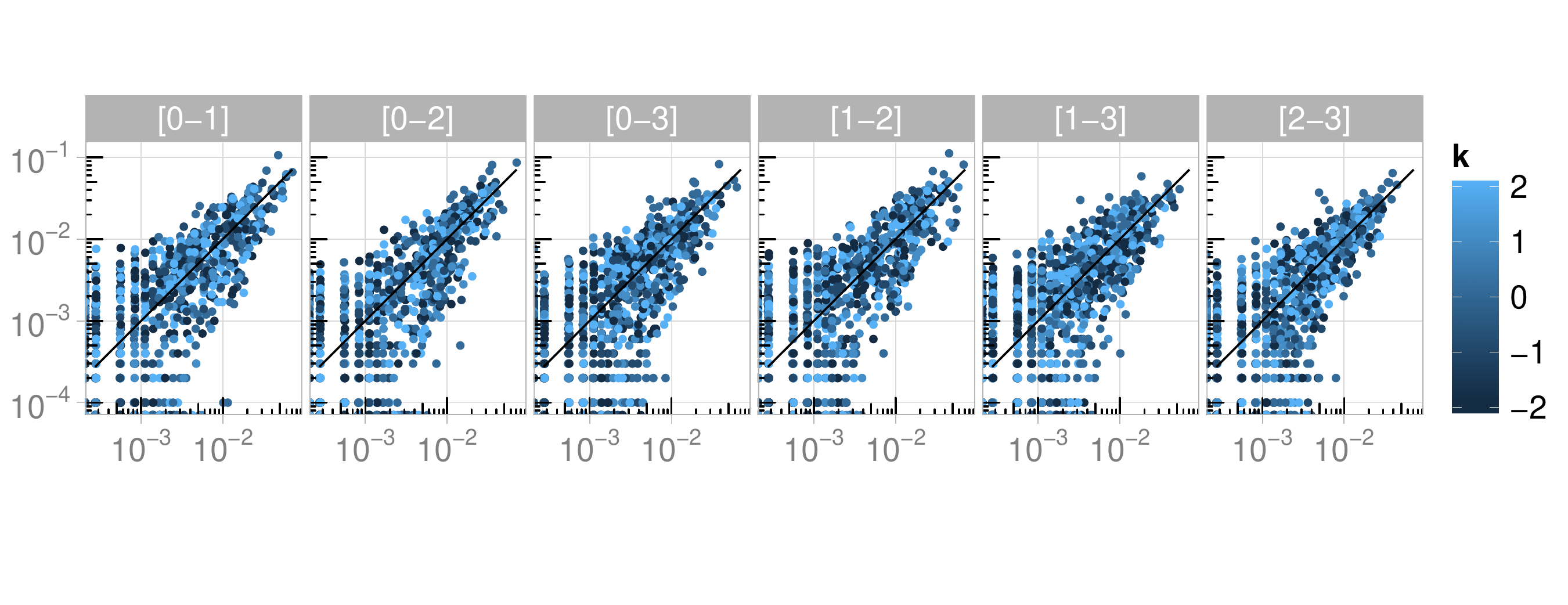}

\vspace{-1cm}

    \caption{Generated correlations versus corpus correlations. We grouped binary connections involving
      the same voices. Colors correspond to the $k$ parameter in
      (\ref{eq:5}). }

  \end{figure*}

We investigated the capabilities of the system to reproduce pair-wise
correlations of the training set. Fig.\ref{modelVSoriginalConnections} shows a scatter plot comparing
the (normalized) values  of each binary connection $f_{ab,ijk}$
in the generated
sequences  versus the ones of the original training corpus. The model
was trained on a  corpus of $51$ major chorales, which
represents the equivalent of a $3244$-beat long chord sequence. We
chose to differentiate horizontal connections from vertical and
diagonal ones by introducing a parameter $L$ as mentioned in
Sect.~\ref{modelDescription}. We took $K=4$, $L=2$,
$\lambda=3e-5$ as parameters and generated a $100000$-beat long sequence. For a discussion on the choice of the regularization
parameter, see Sect.~\ref{parameterChoice}. We see that despite the
small amount of data, the alignment between the generated pair occurrences
and the original ones is quite convincing.

The generation procedure needs solely to compute the ratios
(\ref{eq:2}), which can be done in approximately $O(nK)$
operations. Indeed, since the sequences differ by only one note, only
contributions of its neighboring notes 
has to be taken into account. This has to be compared with the
approximate number (\ref{eq:numPars}) of parameters. Experimentally,
we found that the number of
metropolis steps to achieve convergence is of order
$O(nl\mathcal{A})$ which enables these models to be used in real-time applications.

\begin{figure}
    \centering
  \includegraphics[trim=0 0 0 0, clip, height=5cm]{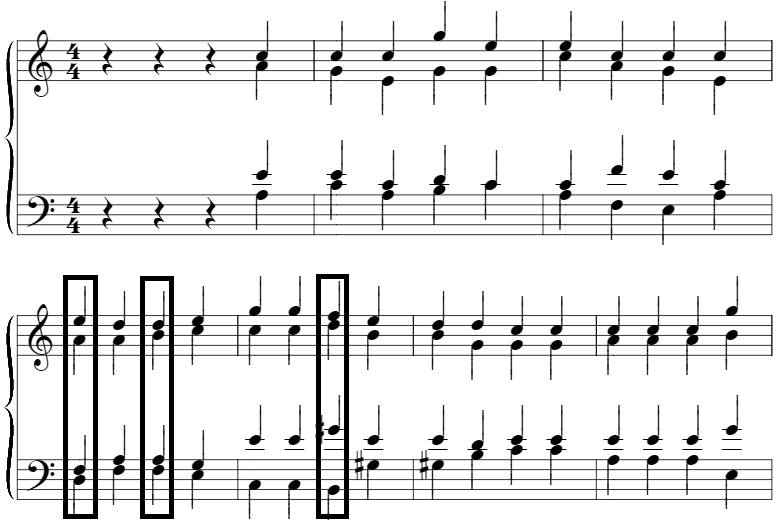}
\caption{Reharmonization of a Bach chorale. ``Invented'' chords are in
boxes. }
  \label{existingChords}
  \end{figure}

We argue that this model does not only reproduce pairwise statistics but can in fact capture  higher-order interactions which makes it suitable for style imitation. Indeed, how the binary connections are combined in Eq.~\ref{eq:6} makes the model able to reproduce correct voicings even if the way  notes composing a chord are distributed among the different voices is a conceptually an interaction of order $n$. Fig.~\ref{existingChords} 
show  chords with nice voicings, voices do not cross, triads have correct doublings and  each separate voice has a coherent shape. A detailed analysis of the chord creation capacity of the system is made in Sec.~\ref{discovery}. We discuss the capability to reproduce other higher-order patterns in Sect.~\ref{sec:highorder}.

\subsection{Chord Invention}
\label{discovery} 
\begin{figure}
    \centering
 \includegraphics[trim=0.2cm 0cm 0cm 0cm, clip, height=4cm]{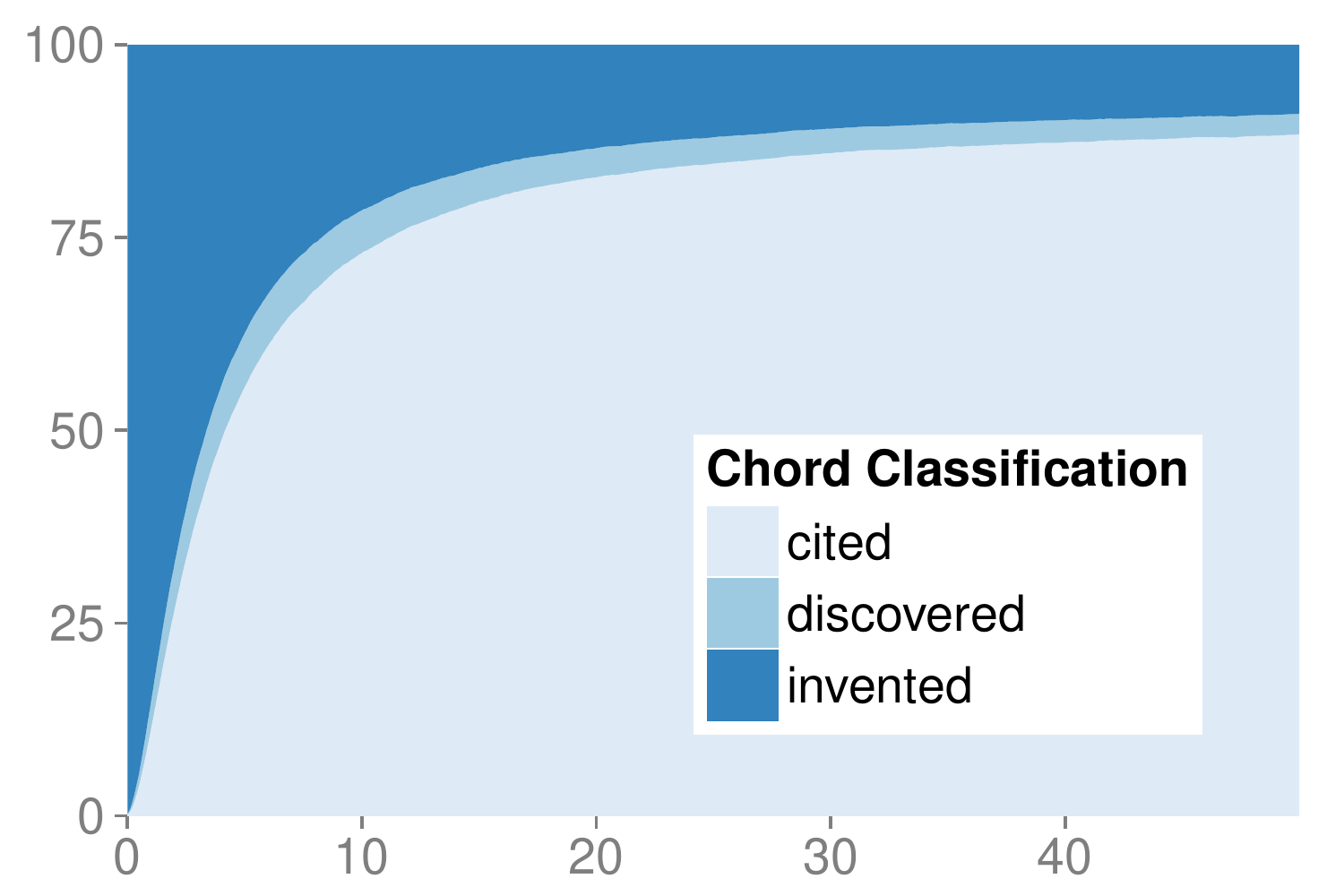}
    \caption{Repartition (in \%) of cited/discovered/invented chords during
      Metropolis-Hastings generation (in normalized number of
      iterations)}
    \label{numberChordsMetropolis}
  \end{figure}

We claim that the competition between the horizontal, vertical and diagonal
correlations of our probabilistic model can generate new chords in the
learned ``style''. Three categories of generated chords can be
distinguished: the \emph{cited} chords which  appear in the model's
training set, the \emph{discovered} chords which do not appear in the
training set but can be found in other Bach's chorales and the
\emph{invented} chords which do not belong to any of the above
categories. 
We used the same model as above to plot in  Fig.~\ref{numberChordsMetropolis} the mean
repartition of chords during the Metropolis-Hastings generation  
as a
function of the number of Metropolis steps (divided by $|\mathcal{A}|nl$). These curves highly depend
on the parameters chosen for the model and on the corpus's size and structure. Nonetheless we
can note the characteristic time for the model to sample from the
equilibrium distribution. For every parameter set we tested, we
observed that when  convergence  is reached, the proportion of
invented and discovered chords seems fixed and significant. 

A closer investigation shows that most of  these ``invented'' chords
 can in fact be classified as
valid in the style of Bach by an expert. The majority of the invented chords is
composed of ``correct'' voicings  of minor or major triads, seventh
chords and chords with nonchord tones.
Fig.~\ref{existingChords} exhibits interesting ``inventions''
 such as an (unprepared) $9-8$ resolution, a dominant ninth and a diminished
seventh. Other invented chords are discordant. A
blindfolded evaluation was conducted to assess to which extent listeners distinguish invented chords from cited ones.
Three non professional music-loving adult listeners were presented with a series of invented chords extracted from generated sequences, and played with their context (i.e. 4 chords before and 4 chords after). 
They were asked whether the central chord was ``good'' or not.
Results show that in average $75\%$ of the invented chords were
considered as acceptable.
\subsection{Higher order interactions}
\label{sec:highorder}
The same analysis as in Sec.~\ref{discovery} can be made for other structures than chords. We chose to investigate to which extent the model is able to reproduce the occurrences of \emph{quadrilateral tuples}. For a sequence $s$, we define the \emph{quadrilateral tuple}  (see Fig.\ref{existingChords}) between voices $i$ and $i'$ at position $j$ to be the tuple
\[(s_{i(j)}, s_{i(j+1)}, s_{i'(j)}, s_{i'(j+1)}).\]

These tuples are of particular interest since many harmonic rules, such as the prohibition of consecutive fiths and consecutive octaves which are often considered to be forbidden in counterpoint, apply to them. Table.~\ref{quad} compares the percentage of cited/discovered/invented generated quadrilateral tuples by the model of Sec.~\ref{imitation}, by a model containing only vertical interactions and by an \emph{independant} model (which reproduces only pitch frequencies).
\begin{table}
\caption{Percentage of cited/discovered/invented quadrilateral tuples}
\centering
\begin{tabular}{lcrrr}
\hline
\noalign{\smallskip}
& \quad & cited & discovered & invented\\
\noalign{\smallskip}
\hline
\noalign{\smallskip}
$K=4, L=2$& \quad & 61.5 &   9.6 &  28.9 \\
$K=0, L=0$ & \quad & 24.1    &  7.8 &  63.1 \\
independant & \quad & 8.4 &   4.4 & 87.2\\
\hline
\end{tabular}
\label{quad}
\end{table}

It shows that an important part of those higher order structures is reproduced. However, analysis exhibits limitations on the higher-order statistics that can be captured (see for instance Fig.~\ref{existingChords}). Indeed, even if our preprocessed corpus contains some of these ``rules violations'', our model is unable to statistically reproduce the number of such structures (they are $2$ to $10$ times more frequent than in the original corpus). We discuss non agnostic methods to integrate these particular rules in Sec.~\ref{discussion}.



\subsection{Flexibility}
\label{sec:reharmonization}
\begin{figure}
    \centering

  \includegraphics[trim=0 90 0 100, clip, width=\textwidth]{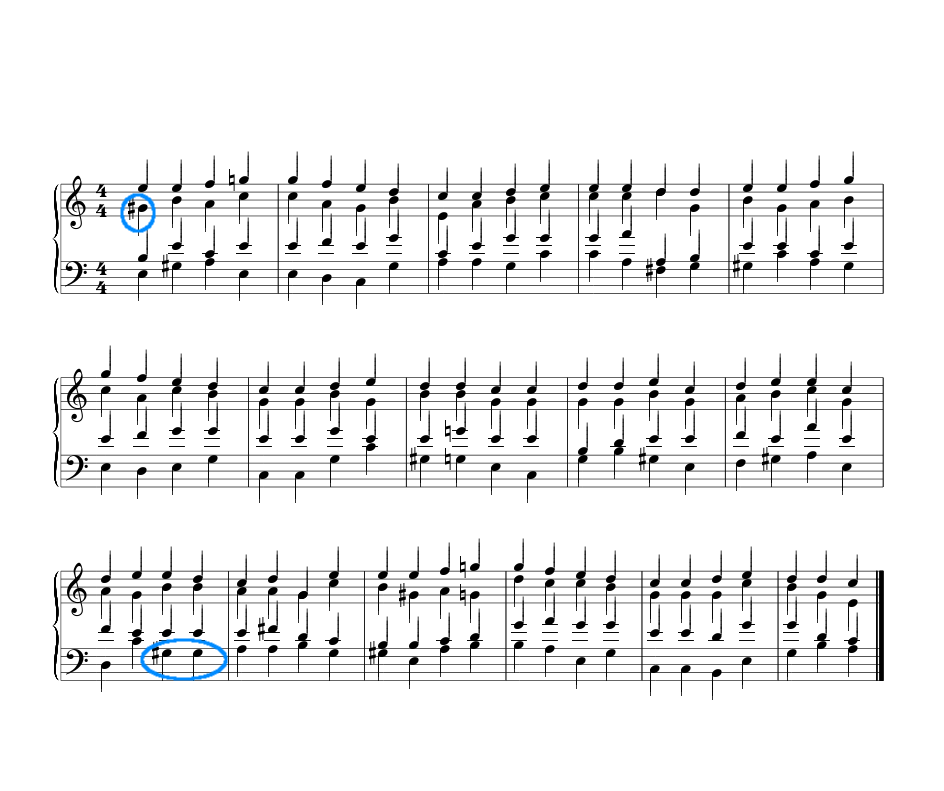}
  \includegraphics[trim=0 90 0 100, clip,  width=\textwidth]{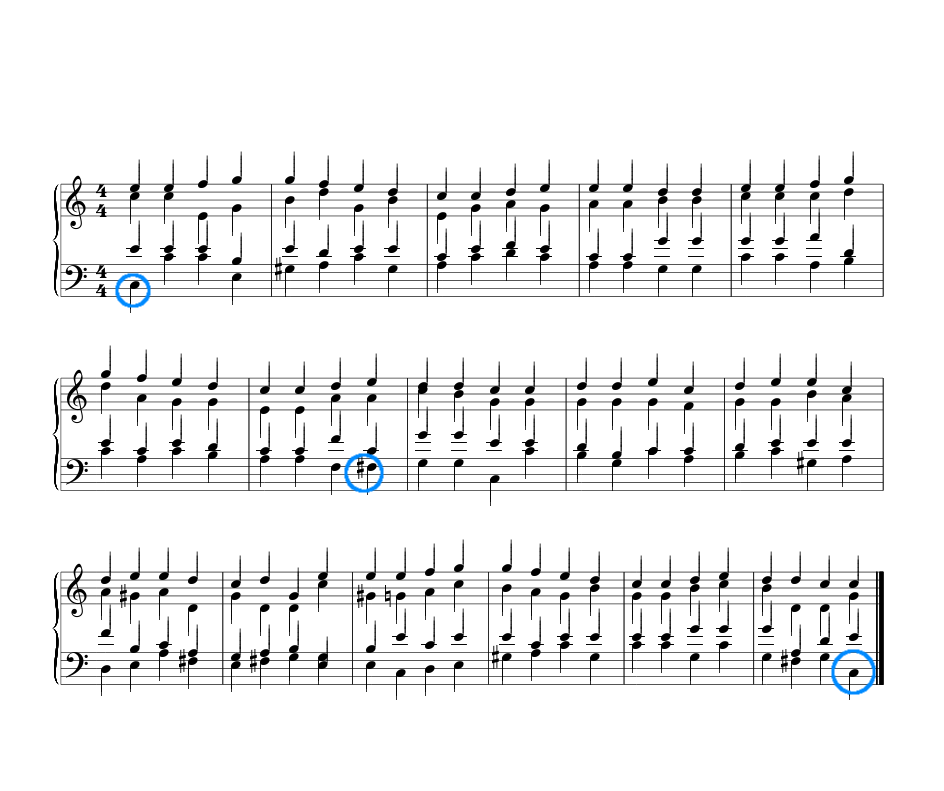}
\caption{Two reharmonizations of Beethoven's Ode to Joy with constraints. Constrained notes are in circles.}

  \label{Ode}
  \end{figure}

As claimed in Sec.~\ref{generation}, we can use our model to generate
new harmonizations of a melody.  Indeed, the simplicity and adaptability of the model allows it to be ``twisted'' in order to enforce unary constraints
while still generating sequences in the learned style. 
As our model is in a specified key (all chorales were transposed
in the same key), we can thus provide convincing Bach-like
harmonizations of plainsong melodies provided they ``fit'' in the
training key. Fig.~\ref{Ode} shows two reharmonizations of Beethoven's Ode to Joy with different unary constraints\footnote{Music examples can
  be heard on the http://flowmachines.jimdo.com/ website}. It is worth noting that even if we put constraints on isolated notes and not on full chords, the constraints propagate well both vertically (the voicings are correct) and horizontally (the progression of chords around the constrained notes is coherent). This opens up a wide range of applications. Those examples show how enforcing simple unary constraints can be used to produce interesting musical phenomena during reharmonization such as:
\begin{itemize}
\item original harmonies (see for instance the $1^{st}$, $2^{nd}$ and $4^{th}$ constraints in Fig.~\ref{Ode})
\item expected harmonies ($3^{rd}$ constraint in Fig.~\ref{Ode})
\item a sense of long-range correlation by creating cadenzas (last constraint in Fig.~\ref{Ode}).
\end{itemize}

By modifying the constrained metropolis
sampling scheme of Sect.~\ref{generation} we can harmonize any
melody $v_0 = [s_{01}, s_{02},  \dots, s_{0l}]$ of size $l$. For each beat $j
\in [1,l]$, we use a melody analyzer to yield the current key $k_j$ at beat
$j$. If the new proposed sequence $s'$ differs from the current one
$s$ by a note at beat $j$, we compute the acceptance ratio
(\ref{eq:2}) by using the probability distribution of the model
trained in key $k_j$. By doing so, we choose the 
appropriate model for each
chunk of the melody and ``glue'' the results together seamlessly.


\subsection{Impact of the regularization parameter}
\label{parameterChoice}
In this section we discuss the choice of the regularization parameter
$\lambda$ of Eq.~\ref{eq:11}. The benefits of introducing
a $L1$-regularization are multiple: it makes  the loss function (\ref{eq:11})
strictly convex (in our case, we do not need to determine if the
family (\ref{eq:4}) is a sufficient family), tends to reduce the
number of non zero parameter coordinates and prevents overfitting 
As our model possesses a
important number of parameters compared to the  number of
samples, adding a regularization term during the training phase
appears to be
mandatory in obtaining good results in the applications we mentioned
Sect.~\ref{generation} and \ref{sec:reharmonization}.

\begin{figure}
    \centering
\includegraphics[height=5cm, width=8cm]{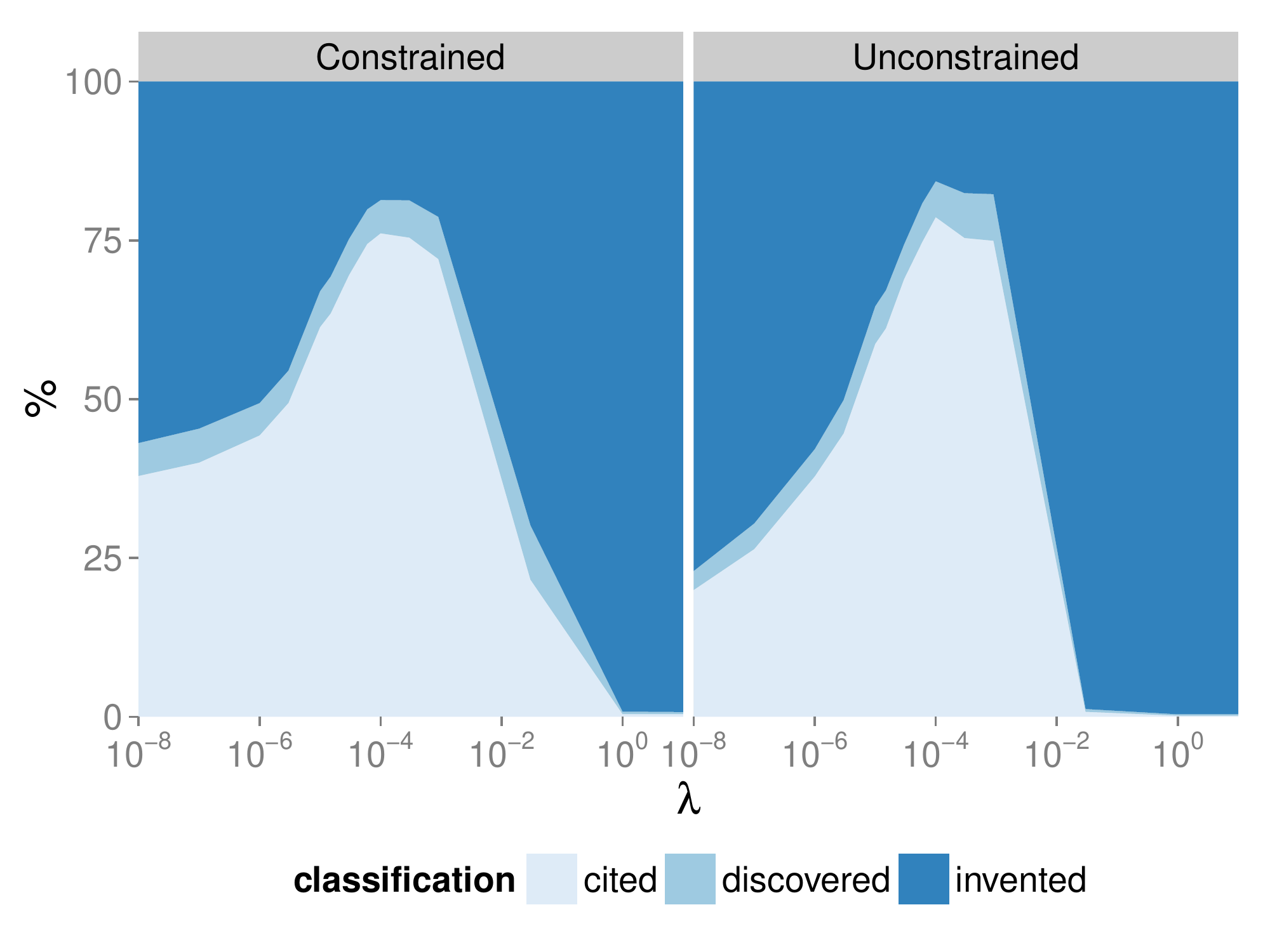}
    \caption{Cited/discovered/invented repartition at equilibrium as a
      function of $\lambda$ in
      the unconstrained and constrained cases.}
 \label{regularizationchoice}
  \end{figure}

We thus evaluate the impact of the choice of $\lambda$ on the
cited/discovered/invented 
classification curves (Fig.~\ref{existingChords}).  We compare the mean repartition of
chords  in the 
unconstrained generation case  and in the
reharmonization case where the first voice is constrained.  The same training corpus as in
Sect.~\ref{generation} is used, with $K=4$, $L=2$ and varying
$\lambda$ . We use the first voice of  chorales from the
testing corpus as constraints. Results are presented in
Fig.~\ref{regularizationchoice}.

A clear influence of $\lambda$ appears for both the unconstrained and constrained
generations. However, these curves are not sharply peaked and it seems not
clear which regularization parameter could be the most
musically-interesting one.

For this, we  investigated to which extent the regularization parameter
influences the model's ability to reproduce a wide variety of chords, either seen in the training set or rediscovered in the testing set.

\begin{figure}
    \centering
\includegraphics[height=5.5cm, width=5.5cm]{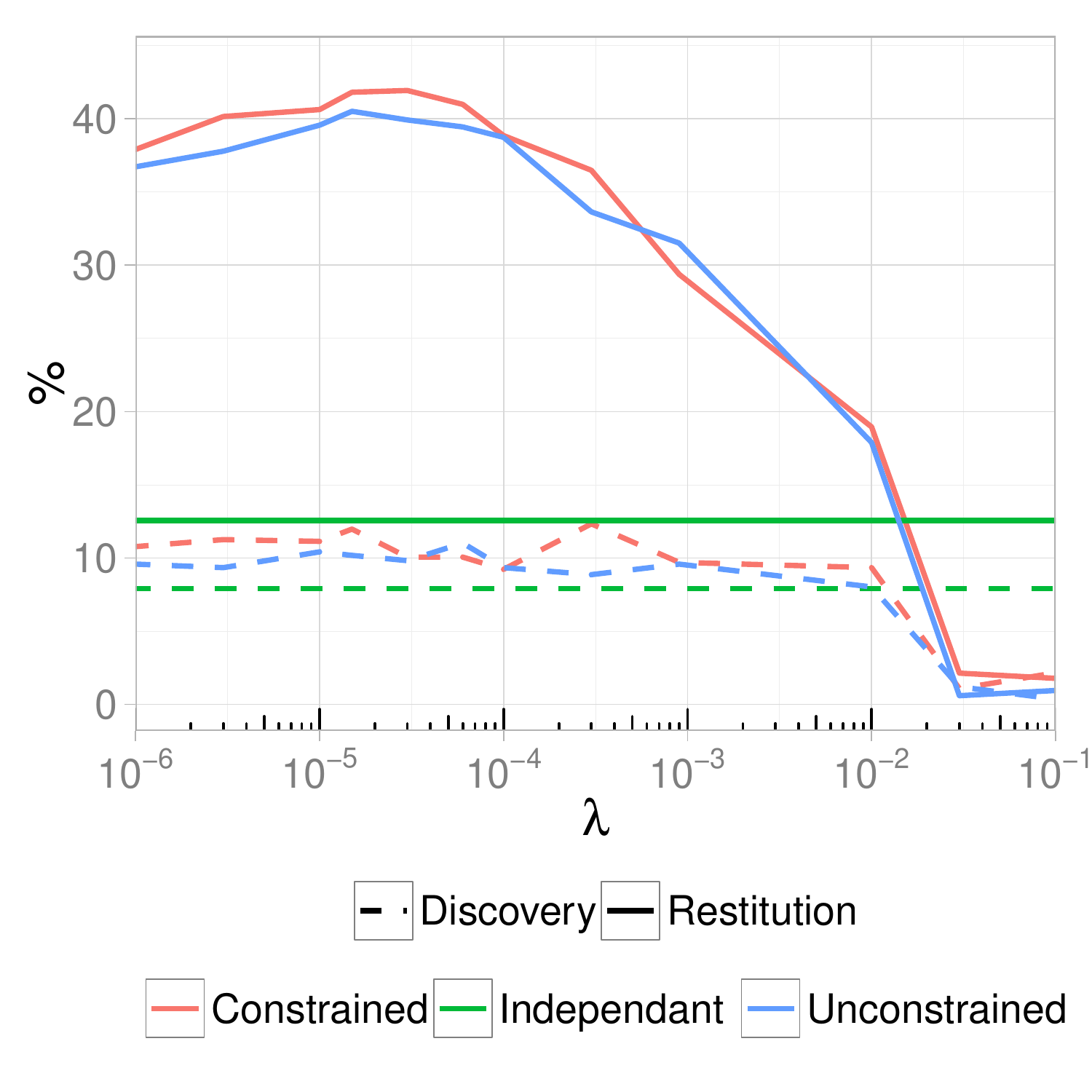}
    \caption{Restitution and discovery percentages as a function of
      $\lambda$ in the unconstrained and constrained cases.}
  \label{uniqueDiscoveries}
  \end{figure}

  
We introduce two quantities revealing the diversity in the generated sequences: the percentage of \emph{restitution} of the training corpus (the number of  cited chords counted without
  repetitions 
normalized by the total number of
  different chords in the training corpus) and
  the percentage of \emph{discovery} of the testing corpus (the number of different discovered chords normalized by the
  number of  chords counted without repetitions in the testing corpus which are not in
  the training corpus).
The evolution of the restitution/discovery percentages as a function
of $\lambda$ is plotted in Fig.~\ref{uniqueDiscoveries} for both the constrained and unconstrained
 generation cases. 
\vspace{-0.5cm}

Both figures exhibit the same behavior. High values of $\lambda$ lead
to uniform models and low values of $\lambda$ to models which
overfit the training data. But the most interesting observation is
that their maximum is not attained for the same value of
$\lambda$ revealing the possibility to control the tradeoff between
the invention capacity, the diversity and the faithfulness with
respect to the training corpus
in the generated sequences.

\subsection{Rhythm}
\label{rhythm}
In this work we focus on chord reproduction and invention. It is an interesting question
since the model is by construction pairwise and chords are, in our examples, four-notes
objects. In order to simplify the analysis on four-notes chords we chose to work on
homorhythmic sequences and we discared all notes not falling on the beat in our corpus,
J.S. Bach's chorales. However, real music is not necessarily homorhythmic. It has rhythm,
\emph{i.e.} notes come in varying durations to form temporal patterns which take place
with respect to some periodic pulse, a kind of temporal canvas. 

We propose a simple
way to extend our model in order to account for rhythmic patterns. 
The model initially presented in \cite{sakellariou:15a} and extended in \ref{model} 
is translation invariant. 
For rhythmic
patterns to emerge we need to break this translation invariance. We do that by introducing
position dependent parameters. More specifically, we choose a cycle which is repeated
over time and within which the translation invariance is broken. Such cycle can be for 
example one or two bars of music. We then divide this cycle in equal time bins which 
correspond to all possible positions where a note can start or end. We call these
time-bins \emph{metrical positions}. We could then define different parameters
between notes on different metrical positions but that would lead to a very large 
number of parameters, yielding a very inaccurate learning (it can be argued that the number of 
parameters must be smaller than the number of data points). We have found that a good compromise
is to let the unary parameters (local fields) be position-dependent while keeping the 
translation invariance for the true binary parameters. This leads to a negligible 
increase in the number of parameters since the unary parameters are of order $|\mathcal{A}|$
whereas the true binary ones are of order $|\mathcal{A}|^2$.
Finaly, in order to obtain
a variety of note durations as well as rests, we introduce two additional
symbols in the alphabet. One symbol for rests and one symbol that signifies the
continuation of the previous note in the current metrical position.

\begin{figure}
    \centering
    \includegraphics[width=0.8\textwidth]{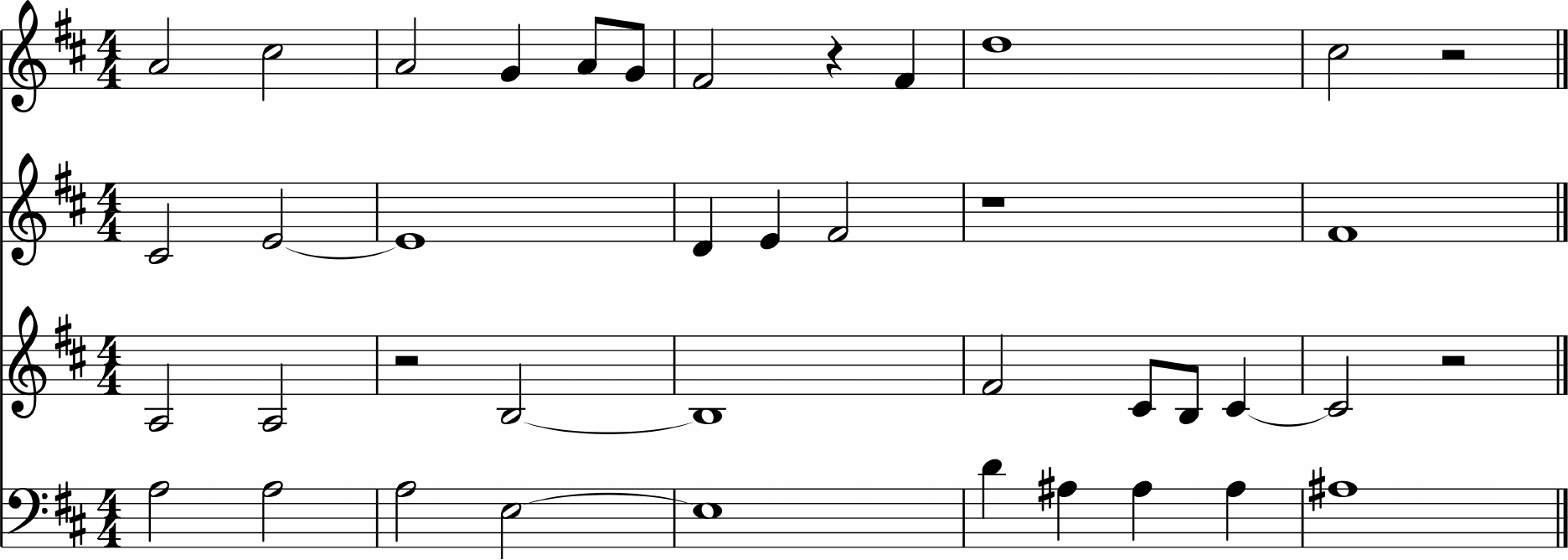}
    \caption{Example of generated polyphonic music with rhythm. The original corpus on which
      the model learned is \emph{Missa Sanctorum Meritis, Credo} by 
      Giovanni Pierluigi da Palestrina}.
\label{palestrina}
\end{figure}

The above procedure has the following effect: the position-dependent parameters
are biasing localy the occurence of symbols (pitches, rests or continuations of the 
previous pitch) in a way that is consistent with the original corpus, which leads
to the emergence of rhythic patterns of the same kind as the ones found in the corpus.
An example can be seen in \ref{palestrina}. For this example we used a cycle of
one bar and divided it in 8 equal parts, corresponding to eighth notes (quavers) which is also 
the smallest division found in the corpus (here \emph{Missa Sanctorum Meritis, Credo} by 
Giovanni Pierluigi da Palestrina). 

\vspace{-0.5cm}

\section{Discussion and future work}
\label{discussion}
We proposed a probabilistic model for chord sequences that captures pairwise dependencies between neighboring notes. 
The model is able to reproduce harmonic
progressions without any prior information and invent new ``stylistically correct'' chords. The possibility to sample with arbitrary unary user-defined constraints makes this model applicable in a wide range of situations. 
We focused mainly on the chord creation and restitution capabilities
which is in our point of view its most interesting feature, an analysis of the plagiarism of monophonic graphical models being made in \cite{sakellariou:15a}.
We showed that even if the original training set is highly combinatorial, these probabilistic methods behave impressively well even if high-order hard constraints such as parallel fifths of octaves cannot be captured.
This method is general and applies to all discrete $n$-tuple sequences. Indeed, we used as features the occurrences of notes, but any other family of functions could be selected.
For instance, adding occurrences of parallel fifths or parallel octaves in the family (\ref{eq:4}) would be possible and would only require $O(n^2|\mathcal{A}|^2)$ parameters, which does not increase our model's complexity.

The utmost importance of the regularization parameter suggests to investigate finer and more problem-dependent regularizations such as group lasso \cite{Friedman2010} or other hierarchical
sparsity-inducing norms  \cite{bach2010}. We believe that having  more than a single scalar regularization parameter $\lambda$ can lead to a better control of the ``creativity'' of our model.

\section*{Acknowledgment}
This research is conducted within the Flow Machines project which received funding from the European Research Council under the European Union’s Seventh Framework Programme (FP/2007-2013) / ERC Grant Agreement n. 291156.


\bibliographystyle{splncs03}
\bibliography{chordcreation}

\end{document}